%% file: MAIN.tex
\documentclass{article} 
\usepackage{colm2024_conference}
\usepackage{microtype}
\usepackage{hyperref}
\usepackage{url}
\usepackage{booktabs}
\usepackage{caption}
\usepackage{subcaption}
\usepackage{tabularray}
\usepackage{multirow}
\usepackage{graphicx}
\usepackage{amsmath}

\usepackage{algorithm}
\usepackage[noend]{algpseudocode}
\usepackage{xcolor}
\usepackage{amsthm}
\newtheorem{definition}{Definition}
\newcommand{\red}{\textbf}
\newcommand{\blue}{\underline}

\title{Exploring prompts to elicit memorization in  masked language model-based named entity recognition}

\author{Yuxi Xia\textsuperscript{1,3}, Anastasiia Sedova\textsuperscript{1,3}, Pedro Henrique Luz de Araujo\textsuperscript{1,3}, Vasiliki Kougia\textsuperscript{1,3}, \\ \textbf{Lisa Nußbaumer\textsuperscript{2}}, \textbf{Benjamin Roth\textsuperscript{1,2}} \\
\textsuperscript{1}Faculty of Computer Science, University of Vienna, Vienna, Austria.\\
\textsuperscript{2}Faculty of Philological and Cultural Studies, University of Vienna, Vienna, Austria.\\
\textsuperscript{3}UniVie Doctoral School Computer Science, Vienna, Austria.\\
\texttt{\{yuxi.xia, benjamin.roth\}@univie.ac.at} \\
}

\begin{document}

\maketitle

\begin{abstract}

Training data memorization in language models impacts model capability (generalization) and safety (privacy risk).
This paper focuses on analyzing prompts' impact on detecting the memorization of 6 masked language model-based named entity recognition models. Specifically, we employ a diverse set of 400 automatically generated prompts, and a pairwise dataset where each pair consists of one person's name from the training set and another name out of the set. A prompt completed with a person's name serves as input for getting the model's confidence in predicting this name. Finally, the prompt performance of detecting model memorization is quantified by the percentage of name pairs for which the model has higher confidence for the name from the training set.
We show that the performance of different prompts varies by as much as 16 percentage points on the same model, and prompt engineering further increases the gap. 
Moreover, our experiments demonstrate that prompt performance is model-dependent but does generalize across different name sets. 
A comprehensive analysis indicates how prompt performance is influenced by prompt properties, contained tokens, and the model's self-attention weights on the prompt.

\end{abstract}
\input{introduction}


\input{related_work}

\input{methodolody}

\input{experiment}

\input{results}

\input{conclusion}
\input{acknowledge}

\bibliography{membership}
\bibliographystyle{colm2024_conference}

\appendix
\section{Appendix}

\subsection{Confidence of \textit{PER} in a Prompt} \label{confidence}
We describe the details of obtaining the confidence of a $PER$ in a prompt in Algorithm \ref{euclid}.
\begin{algorithm}
    \footnotesize
    \caption{Confidence of $PER$ in a Prompt}\label{euclid}
    \hspace*{\algorithmicindent} \textbf{Input:} a person's name $PER$, a prompt $PT_k$, a NER model $M_{ner}$, tokenized $PER$ set $T_{PER}$, token set $T_{PT_k \wedge PER}$ after tokenizing $PT_k \wedge PER$. 
    
    \begin{algorithmic}[1]
    \State $C_k(PER) \leftarrow 0$
    \State $Pr(T^1), Pr(T^2), ..., Pr(T^{|T_{PT_k \wedge PER}|}) \leftarrow M_{ner}(T_{PT_k \wedge PER})$.
    \For {$t \leftarrow 1$ to $|T_{PT_k \wedge PER}| $}
        \If {$T^t \in T_{PER}$}
            \State $C_k(T^t) \leftarrow max(Pr(\textit{B-PER} | T^t), Pr(\textit{I-PER} | T^t))$
            \State $C_k(PER) \leftarrow C_k(PER) + C_k(T^t)$
        \EndIf
    \EndFor
    \State $C_k(PER) \leftarrow C_k(PER)/|T_{PER}|$.
    \State  \textbf{Output: } $C_k(PER)$, confidence score of $PER$ for $PT_k$
    \end{algorithmic} \label{a:confidence}
\end{algorithm}

\subsection{MLM-based NER Models} \label{app:models}
We summarize the model performance details of the 6 explored fine-tuned NER models based on MLMs in Table \ref{tab:models}, and more details about the models and accessibility information are described in the following.

\textbf{ALBERT-B \footnote{\url{https://huggingface.co/ArBert/albert-base-v2-finetuned-ner}} and ALBERT-L \footnote{\url{https://huggingface.co/Gladiator/albert-large-v2_ner_conll2003}}} NER models are fine-tuned models of base and large versions of ALBERT-v2 \citep{lan2020albert} respectively. ALBERT-v2 is a transformer-based language model that is pre-trained on a substantial corpus of English text data through self-supervised learning. This pretraining process entails exposure to raw textual data without human annotation, thereby leveraging publicly available data sources. Specifically, ALBERT-v2 utilizes two pretraining objectives: Masked Language Modeling (MLM) and Sentence Ordering Prediction (SOP). Through MLM, the model learns to predict masked tokens within a sequence of text, while SOP tasks involve predicting the correct order of sentences within a document. Notably, ALBERT-v2 aims to optimize efficiency and scalability by parameter sharing across layers and employing parameter reduction techniques, such as factorized embedding parameterization and cross-layer parameter sharing.

\textbf{BERT-B \footnote{\url{https://huggingface.co/dslim/bert-base-NER}} and BERT-L \footnote{\url{https://huggingface.co/dslim/bert-large-NER}}} NER models have the most downloaded records on the public model platform. These two models are fine-tuned models of  BERT-base-cased and BERT-large-cased models \citep{devlin2019bert}. BERT is a transformer-based language model renowned for its ability to capture bidirectional contextual information from text data. Similar to ALBERT, BERT models are pre-trained on a large corpus of English data in a self-supervised fashion, aiming to learn deep contextualized representations of words or tokens. BERT employs the MLM objective, where it learns to predict masked tokens within a sequence, and the Next Sentence Prediction (NSP) objective, where it predicts whether two sentences are consecutive in the original text. BERT's architecture comprises transformer encoders stacked on top of each other, allowing it to effectively capture contextual information through attention mechanisms.

\textbf{RoBERTa-B \footnote{\url{https://huggingface.co/dominiqueblok/roberta-base-finetuned-ner}} and RoBERTa-L \footnote{\url{https://huggingface.co/Gladiator/roberta-large_ner_conll2003}}} NER models  are fine-tuned models of RoBERTa-base and RoBERTa-large \citep{liu2019roberta}. RoBERTa is another variant of the BERT model designed to improve upon its pretraining methodology and performance. Like BERT and ALBERT-v2, RoBERTa utilizes transformer-based architectures for language modeling tasks. However, RoBERTa introduces several enhancements to the pretraining process, including dynamic masking strategies, larger training datasets, and longer training times. Notably, RoBERTa replaces the NSP objective with a more extensive masking scheme during pretraining to improve the model's robustness and effectiveness in capturing contextual information from text data. Additionally, RoBERTa employs larger batch sizes and longer training sequences, contributing to its improved performance on downstream natural language processing tasks.

\begin{table}
    \centering
    \begin{tabular}{lcccccc}
    \hline
        Model &\#Param.& P & R  & F1 & ACC & Training epoch\\\hline
        
        ALBERT-B & 11.1M& 93.0 &93.8  &93.4 & 98.5 &3 \\
        ALBERT-L & 16.6M& 94.0 &94.5  &94.2 & 98.7 &5\\
        BERT-B & 108M& 92.1 & 93.1 & 92.6&91.2 &-\\

        BERT-L &  334M& 92.0 & 91.9 &92.0&90.3 & - \\
         Roberta-B &  124M& 95.3&96.0  &95.7 &98.9 &3\\
        Roberta-L &  354M&  96.2 &96.9  &96.6 & 99.4 & 5 \\
        \hline
    \end{tabular}
    \caption{Self-reported performance of the NER Models on CoNLL-2003 evaluation dataset.}
    \label{tab:models}
\end{table}

\input{appendix_token_level_analysis}

\subsection{Self-attention Analysis}\label{other_heatmaps}

\begin{figure}[ht!]
\begin{subfigure}[t]{.95\linewidth}
\includegraphics[width=\textwidth]{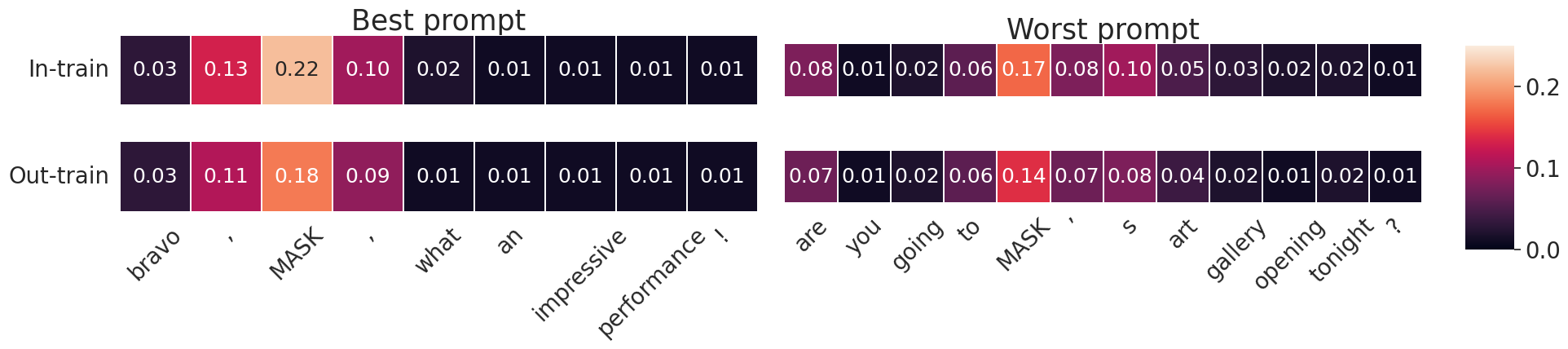}
   \caption{ALBERT-B}
\end{subfigure}
\hfill
\begin{subfigure}[t]{.95\linewidth}
\includegraphics[width=\linewidth]{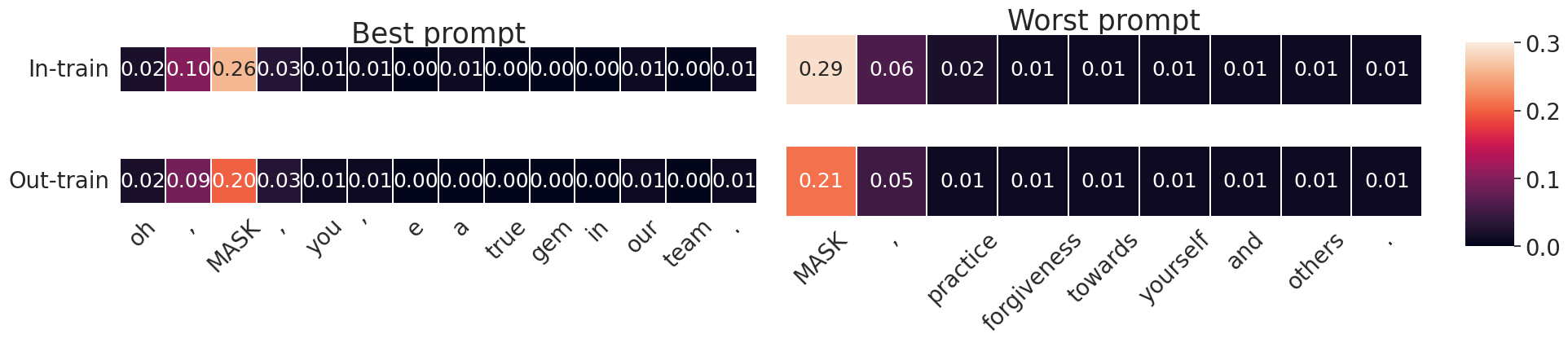}
\caption{ALBERT-L}
\end{subfigure}
\hfill
\begin{subfigure}[t]{.95\linewidth}
\includegraphics[width=\linewidth]{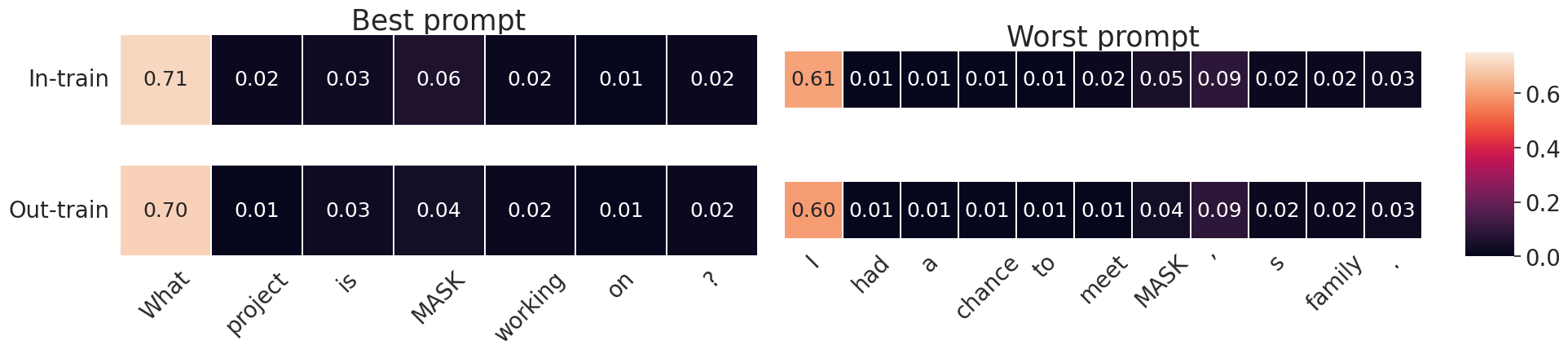}
\caption{BERT-L}
\end{subfigure}
\hfill
\begin{subfigure}[t]{.95\linewidth}
\includegraphics[width=\linewidth]{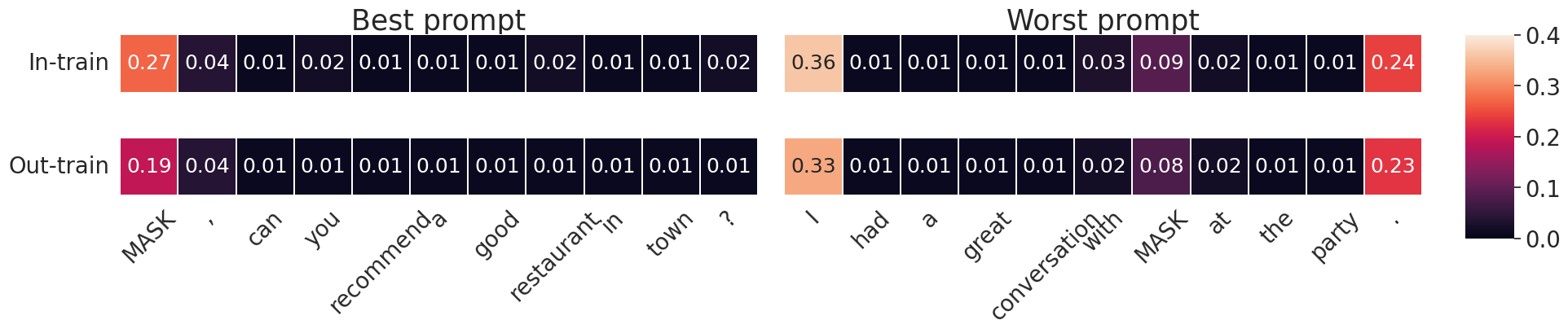}
\caption{RoBERTa-L}
\end{subfigure}
\hfill
\begin{subfigure}[t]{.95\linewidth}
\includegraphics[width=\linewidth]{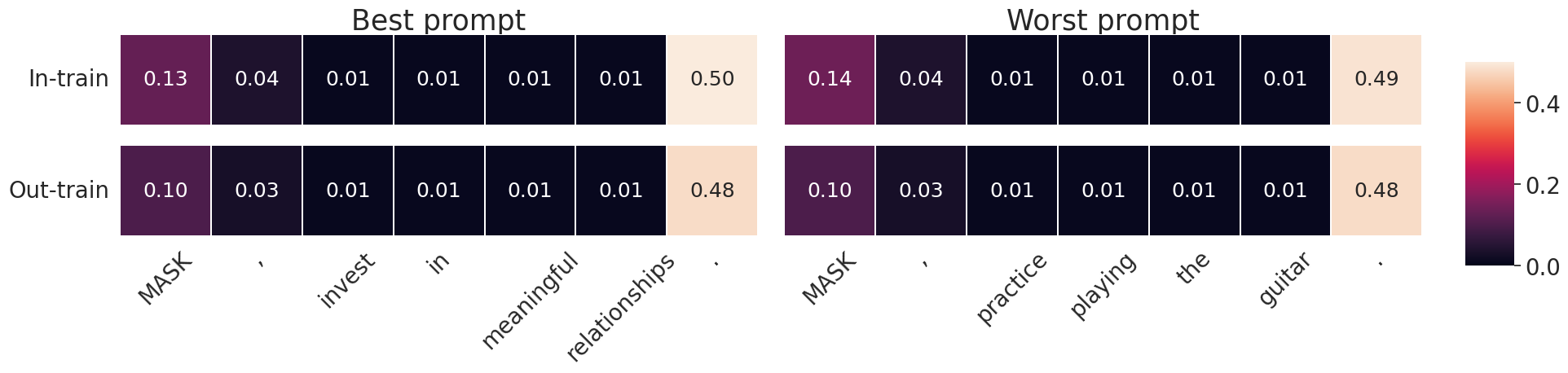}
\caption{RoBERTa-L}
\end{subfigure}
    \caption{Attention heatmaps of the best and the worst prompts across 5 models on the dev set averaged over all attention heads and layers. The attention weights corresponding to each prompt's ``MASK'' token are used and averaged over the In-train and Out-train $PER$ sets separately.}
    \label{fig:all_heatmaps}
\end{figure}

Same as Section \ref{alyn: self-attention}, we provide self-attention analysis for the rest of the 5 models in Figure \ref{fig:all_heatmaps}.

\end{document}

%% file: introduction.tex
\section{Introduction}

Recent studies have highlighted the memorization of training data in language models \citep{neel2023privacy}, especially for auto-regressive models \citep{mireshghallah2022empirical, carlini2023quantifying}. Such studies are important in exploring model capability (generalization \citep{feldman2020does}) or safety (privacy risk \citep{neel2023privacy}). However, few works \citep{ali_unintended_2022} explore memorization in Masked Language Model (MLM)-based Named Entity Recognition (NER) models, despite their state-of-the-art performances in different domains, including health records \citep{richie2023extracting}, social media \citep{yu-etal-2023-grounded}, and legal documents \citep{darji2023german}. 


Unlike \citet{ali_unintended_2022}, who use only 5 hand-written prompts to detect model memorization of a self-fine-tuned NER model on their private training data (not publicly accessible), we consider the sensitivity of model confidence towards individual prompt variations \citep{feng2024unveiling}, and thus employ a set of 400 diverse prompts generated by a generative language model (i.e., ChatGPT \citep{OpenAI_ChatGPT_2023}). The prompt set covers 4 types of prompts (i.e., declarative, exclamatory, imperative, and interrogative), 15 prompt token lengths, and 10 token positions for holding the target entity (person's name). We focus on exploring the prompts’ impact on detecting the memorization of persons' names in 6 publicly accessible NER models fine-tuned on the CoNLL-2003 dataset \citep{tjong-kim-sang-de-meulder-2003-introduction}. 

To quantify model memorization, we create a 
pairwise dataset sampled from Wikidata \citep{vrandevcic2014wikidata} consisting of a large number ($826^2+825^2$) of person's name (\textit{PER}) pairs. This dataset is divided into development (dev) and test sets, each containing different named entities (\textit{PER}s). Every instance in both sets is a pair of \textit{PER}s, composed of an In-train \textit{PER} (from the training data of the NER model) and an Out-train \textit{PER} (not present in the training data). 

\begin{figure*}
  \centering
  \includegraphics[width=0.99\linewidth]{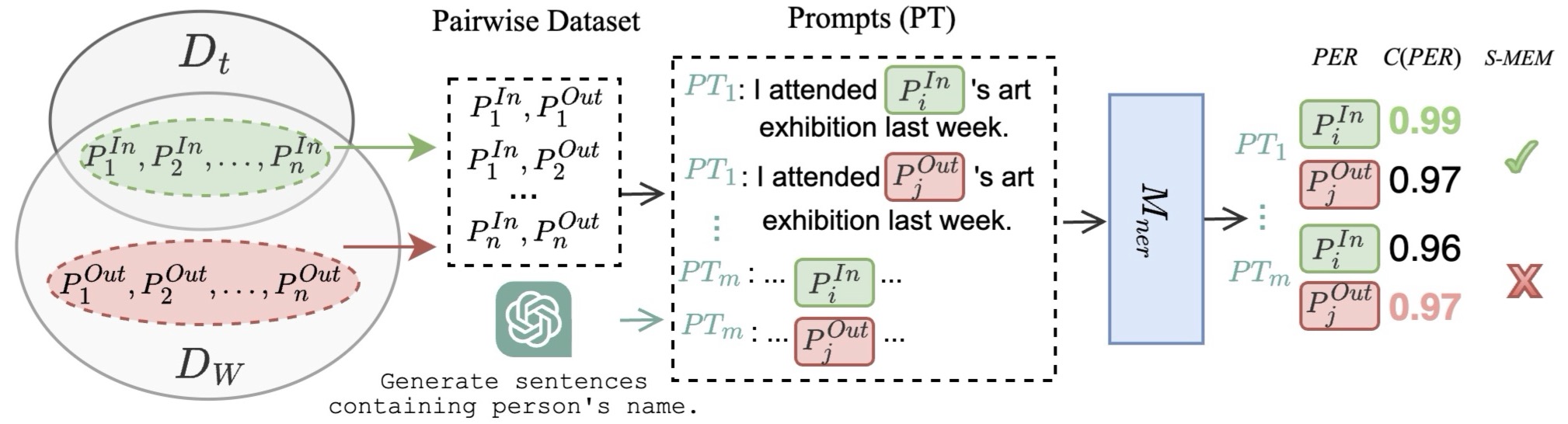}
  \caption{Memorization detection of a NER model using diverse prompts and the pairwise dataset. $P^{In}_*$ and $P^{Out}_*$ refer to a person's name in and out of the training data respectively.}
  \label{fig:framework}
\end{figure*}

Figure \ref{fig:framework} illustrates the full pipeline for model memorization detection. Specifically, each prompt from the prompt set is completed separately with all \textit{PER} entities from the pairwise dataset. A completed sentence, consisting of a prompt and a $PER$, serves as input to a NER model to get the confidence score of the $PER$ in this prompt. We define that given an In-train and Out-train \textit{PER} pair, the pair exposes sample memorization (\textit{S-MEM}) if the In-train \textit{PER} receives a higher confidence score than the Out-train \textit{PER}. Ultimately, the prompt's performance in detecting model memorization is quantified by the percentage of name pairs that expose sample memorization in the pairwise dataset.


Overall, our contributions are: 

\begin{itemize}
    \item We are the first to analyze the prompt's influence on the memorization detection of MLM-based NER models, using 400 diverse prompts and 6 publicly accessible NER models, and we study their impact across different model properties. 
        
    \item We show that memorization detection is sensitive to the prompt choice. Performance differences of prompts in the set on the same model can be as large as 16 percentage points. 
    Prompt performance is model-dependent but generalizes across mutually exclusive name sets (from dev to test data).    
  
    \item We apply ensembling and prompt engineering techniques on the prompt set and observe that the engineered (modified) prompts can improve performance by up to 2 percentage points compared to the best-performed prompt in the prompt set.
    
    \item We comprehensively analyze how prompt performance is influenced by various factors, including the properties of the prompt, the tokens it contains, and the self-attention weights of the model focused on the prompt.

    
\end{itemize}

%% file: related_work.tex
\section{Related Work} 


\textbf{Memorization in auto-regressive language models} was first studied by \citet{carlini2019secret} who showed that LSTMs memorize a significant fraction of their training data. Later, \citet{carlini2021extracting} and \citet{carlini2023quantifying} use extractability to measure the memorization of GPT-2 and GPT-Neo models \citealt{wei2023overview}, showing that prefixed prompts can extract sensitive training data. However, extractability only applies to auto-regressive language models that use only the input side of fine-tuning data, but not to fine-tuned MLMs that use also the output side (input labels). Several studies \citep{feldman2020does, brown2021memorization, razeghi2022impact} indicate that a certain amount of memorization is not necessarily undesirable but helpful for model generalization. 
 \citet{tirumala2022memorization} and \citet{magar2022data} study \textbf{memorization in MLMs} towards the model generalization using a metric as same as accuracy and accuracy difference between training and test sets respectively. \citet{ali_unintended_2022} focus on the \textbf{memorization in fine-tuned MLMs} (NER models) and average the confidence ranks of In-train entities in an entity set to measure model memorization. However, \citet{ali_unintended_2022} self-fine-tune the NER model and study the impact of the training data duplication and complexity on the memorization dynamics.  Differently, we explore 6 publicly accessible NER models fine-tuned by different developers and focus on the prompts' impact on the memorization detection of the NER models. 

\textbf{Hand-written vs Generated Prompts.} Language model performance has been shown to be sensitive to prompt variations in previous works (\citealt{feng2024unveiling, sclar2023quantifying}). 
Thus, randomly choosing one of the 5 hand-written prompts for each studied entity to detect memorization as in \citet{ali_unintended_2022} may not accurately represent the model memorization. 
Instead, we study \textbf{400 generated diverse prompts}, showing the big variances of memorization values detected by different prompts individually were not considered in previous settings.

%% file: methodolody.tex
\section{Methodology}

In this section, we first describe the processes for creating the pairwise dataset and prompt set. Then, we introduce how to use them to detect memorization in NER models.
\subsection{Pairwise Dataset Creation}

We create a pairwise dataset $D_{pw}$ to quantify the memorization of NER models fine-tuned on the CoNLL-2003 dataset \citep{tjong-kim-sang-de-meulder-2003-introduction}. Specifically, we first generate datasets $D_t$ and $D_W$ which contain all $PER$s from the CoNLL-2003 dataset and Wikidata \citep{vrandevcic2014wikidata} respectively. $D_{pw}$ is then sampled from the $PER$s in $D_W$ considering the $PER$s in $D_t$. The details of the datasets are shown in Table \ref{tab:dataset}: 

\textbf{NER-train $PER$ Dataset ($D_t$).} CoNLL-2003 is a commonly used NER dataset with a large number of examples of named entities. The training dataset of the CoNLL-2003 corpus contains 6,600 $PER$s, i.e., entities with ground-truth label \textit{B-PER} and/or \textit{I-PER} which denote the beginning and the rest of a person’s name respectively. After the post-processing of removing the duplicates and single-token $PER$, $D_t$ contains 2,645 unique multi-token $PER$s.


\textbf{Wikidata $PER$ Dataset ($D_W$).} Wikidata as a source for $D_W$ enables the compilation of a list of real-world person names.
Specifically, a SPARQL query is formulated to retrieve $PER$ entities. 
Similar to $D_t$, duplicates and single-token $PER$ are filtered out. Finally, 7,617,797 multi-token $PER$ form the dataset $D_W$.

\textbf{Pairwise Dataset ($D_{pw}$).} We only sample the $D_{pw}$ from $D_W$ to ensure same data source for both In-\&Out-train $PER$s. The intersection of $D_W$ and $D_t$ ($D_W \cap D_t$) contains 1,651 $PER$s, which onstitutes to the In-train $PER$ set $\{P^{In}_i\}^{N}_{i=1}$, \textit{N=}1,651. In contrast, an equal size of \textit{Out-train} $PER$ set $\{P^{Out}_j\}^{N}_{j=1}$ are sampled from $D_W$ excluding $D_t$ ($D_W\setminus D_t$). As there is no training process in our setting, we split $\{P^{In}_i\}^{N}_{i=1}$ and $\{P^{Out}_j\}^{N}_{j=1}$ sets equally into development (dev) and test respectively. In the end, the dev data is constructed with $\{P^{In}_i\}^{n}_{i=1}$ and $\{P^{Out}_j\}^{n}_{j=1}$, $n=826 (\approx N/2)$, the rest of the $PER$s goes to the test data. The test data is used to verify the generalization of prompt performance across different name sets (from dev to test). Each instance in dev and test sets is a name pair consisting of a \textit{In-train} $PER$ and a \textit{Out-train} $PER$, and every $PER$ in $\{P^{In}_i\}^{n}_{i=1}$ is paired with all the $PER$ in $\{P^{Out}_j\}^{n}_{j=1}$, resulting in $n^2$ instances in the dev data of $D_{pw}$.

\input{tables/data_descrip}

\subsection{Prompt Set Generation}

To collect a large and diverse set of prompts, we use a generative large language model, GPT-3.5-turbo-1106 \citep{OpenAI_ChatGPT_2023} (ChatGPT), to automatically generate $m$ ($m=400$) diverse prompts $\{PT_k\}_{k=1}^m$. 
Specifically, to increase the diversity of the prompts, we consider four categories of sentences: (1) declarative; (2) exclamatory; (3) imperative; and (4) interrogative. 
Next, we prompted ChatGPT to generate 100 sentences for each category. To collect exclamatory prompts, we use the following prompt:  
\textit{``Generate 100 different exclamatory sentences that must contain persons' names, and replace the person's name with 'MASK'''}. The prompts of other categories are generated by replacing the \textit{``exclamatory''} string.
The \textit{``MASK''} string in a prompt $PT_k$ is completed with the $PER$ from the pairwise dataset, and the completed sentence $PT_k\wedge PER$ is fed to NER models to obtain the confidence score of the $PER$ entity in $PT_k$.

\subsection{Memorization Detection}


Let $M_{ner}$ denote a NER model that has been fine-tuned on a pre-trained MLM with dataset $D_t$. Such models can achieve high classification accuracy (shown in Table \ref{tab:models} in the Appendix \ref{app:models}) in predicting entity labels like $PER$ (i.e., 
\textit{B-PER} and \textit{I-PER} labels). 
To compute the confidence score of a $PER$, we first gather the highest likelihood scores between \textit{B-PER} and \textit{I-PER} labels for all name tokens of the $PER$, and then take the mean value of likelihood scores as the final confidence for the $PER$. The confidence of a $PER$ in $PT_k$ is defined as the notation $C_k(PER)$ (more details show in Algorithm \ref{a:confidence} in the Appendix \ref{confidence}).
After getting the confidence score for the In-train and Out-train $PER$s in a name pair, we identify the sample memorization of this $PER$ pair using the following definition:
\begin{definition}
\textbf{Sample Memorization (\textit{S-MEM})}: Given an In-train and Out-train \textit{PER} pair, the pair exposes the sample memorization (S-MEM) of a NER model if the In-train \textit{PER} gets a higher model confidence score than the Out-train \textit{PER}.
\end{definition}

The sample memorization of a $PER$ pair in the prompt $PT_k$ can be formulated as: 
$$
\textit{S-MEM}_{M_{ner}} (PT_k \wedge P^{In}_i, PT_k \wedge P^{Out}_j) = 
\begin{cases}
  1  & \text{if } C_k(P^{In}_i) > C_k(P^{Out}_j) \\
  0  & \text{if } C_k(P^{Out}_j) > C_k(P^{In}_i)
\end{cases}$$


To detect the memorization of a NER model, we further define:

\begin{definition}
    \textbf{Model Memorization (\textit{M-MEM})}: The memorization of a NER model is quantified by the percentage of name pairs that expose sample memorization in a pairwise dataset.
\end{definition}


Finally, we formulate the performance of a prompt $PT_k$ on detecting memorization of $M_{ner}$ as \textit{M-MEM} scores: 
$\textit{M-MEM}_{M_{ner}}(PT_k) = \sum_{i=1 }^n \sum_{j=1 }^n \textit{S-MEM}_{M_{ner}}  (PT_k \wedge P^{In}_i, PT_k \wedge P^{Out}_j) / n^2$.






        

%% file: tables/data_descrip.tex
\begin{table}
    \centering
    \footnotesize
    \begin{tabular}{ccccc}
    \hline
        Dataset & Source data & \#\textit{PER}  & \#Instance & Property \\\hline
        $D_t$ & CoNLL-2003 & 2,645 & 2,645 & \textit{In-train }  \\
        $D_W$&  Wikidata & 7,617,797 & 7,617,797  &\textit{In-} \& \textit{Out-train} \\
        $D_{pw}$ & $ D_W \cap D_t $  + $D_W\setminus D_t$   & 1,651 + 1,651  & $826^2$(dev) + $825^2$(test) & \textit{In-} \& \textit{Out-train}\\
         \hline
    \end{tabular}
    \caption{Datasets details. }
    \label{tab:dataset}
\end{table}

%% file: experiment.tex
\section{Experimental Setup}

We performed our model memorization analysis on publicly accessible MLM-based NER models fine-tuned by different developers on the CoNLL-2003 dataset \citep{tjong-kim-sang-de-meulder-2003-introduction}.
To provide a more comprehensive analysis, we selected 6 models built on MLMs across 3 different pretraining schemes (ALBERT-v2 \citep{lan2020albert}, BERT \citep{devlin2019bert}, and RoBERTa \citep{liu2019roberta}) and 2 model sizes (base and large sizes, \textit{-B} and \textit{-L} notations are used correspondingly) for each scheme. 
More details about the models and accessibility information can be found in Appendix \ref{app:models}. 

\subsection{Strategies of Using the Prompt Set} 

We investigate three strategies for exploiting the prompt set, aiming for a more thorough exploration of the impact of prompts on the memorization detection of NER models.

\textbf{Baselines (BS).} \textbf{$\emptyset$-PT} uses the $PER$ without any additional text as the input to query corresponding confidences for memorization detection. \textbf{One-PT} and \textbf{Mix-PT} use the same strategies in \citet{ali_unintended_2022}, \textbf{One-PT} employs one hand-written prompt (\textit{``My name is XX.''}), corresponding to the \emph{known}-setting in \citet{ali_unintended_2022}, while \textbf{Mix-PT} randomly chooses for each name one of the 5 hand-written prompts (\textit{``My name is XX.''}, \textit{``I am XX.''}, \textit{``I am named XX.''}, \textit{``Here is my name: XX.''}, \textit{``Call me XX.''}), like the \emph{unknown}-setting in \citet{ali_unintended_2022}.

\textbf{Original Prompt (OPT)} selects the best prompt (\textbf{B-PT}) and worst prompt (\textbf{W-PT}) from the prompt set that achieved the highest and lowest \textit{M-MEM} scores on the dev set of $D_{pw}$.

\textbf{Prompt Engineering (PTE)} is inspired by \citet{feng-etal-2018-pathologies} which applies a token-removal operation to the originally generated prompts.
The goal is to maximize the \textit{M-MEM} score gap of the original best and worst prompts. 
Specifically, the most important token (which contributes the most to score improvement) and the least important token are removed from the worst and best prompts, respectively, resulting in two modified prompts. 
This process is iteratively repeated for the modified prompts until only one token (excluding the \textit{PER}) remains.
The two modified prompts that achieved the highest and lowest \textit{M-MEM} scores on the dev set among all the modified prompts are named \textbf{BM-PT} (best-modified prompt) and \textbf{WM-PT} (worst-modified prompt). 
It is important to note that such prompts may be ungrammatical.
More details are presented in Section~\ref{aly:token}.

\textbf{Ensembling of Prompts (EPT)} employs several ensemble techniques: (1) Majority Voting (\textbf{MV}) decide the \textit{S-MEM} of a name pair if more than half of the prompts in the set vote ``1''; (2) Average Confidence Score (\textbf{AVG-C}) uses the average $PER$ confidence over all prompts ($avg(\{C_k(PER)\}_{k=1}^{k=m}$)) as the final confidence score of the $PER$; (3) Weighted Confidence Score (\textbf{WED-C}) weights the $PER$ confidence of each prompt by its \textit{M-MEM} score; (4) Maximum confidence score (\textbf{MAX-C}) denote the maximum confidence of the $PER$ over all prompts ($max(\{C_k(PER)\}_{k=1}^{k=m}$)) as the final confidence score; (5) Minimum confidence score (\textbf{MIN-C}) is the contrast to MAX-C, which uses the minimum value ($min(\{C_k(PER)\}_{k=1}^{k=m}$)) as the final confidence.

%% file: results.tex
\input{tables/best_worst_prompt}

\section{Results and Analysis}

Table~\ref{tab:bestWorstPrompts} shows each model's highest and lowest \textit{M-MEM} scores achieved on the dev set by the best and worst prompts in the original prompt set respectively.
The \textit{M-MEM} scores vary considerably depending on the prompt: the gap between the best and the worst prompt for a given model is as large as 16.40 percentage points for ALBERT-B on the test set.
We validate the statistical significance of the results through Cochran's  $\mathcal{Q}$ test. 
We find that the differences are significant for all models ($\text{p}<0.001$). The following sections summarize and analyze the results to answer 6 research questions. 

\subsection{Do Model Properties Affect the Memorization Scores?}
The 6 investigated models mainly cover 3 different pretraining schemes (ALBERT-v2, BERT and RoBERTa) and 2 different model sizes for each scheme. There is no \textbf{pretraining scheme} led to higher/lower memorization from results in Table ~\ref{tab:bestWorstPrompts} and Figure~\ref{fig:promptRankCorrs} (a). However, we observe that the score gaps of large models are generally smaller than those of base models in Table ~\ref{tab:bestWorstPrompts}, this trend does not apply to the standard deviations of the \textit{M-MEM} scores of all prompts, as shown by Figure~\ref{fig:promptRankCorrs} (a). For example, a similar standard deviation is observed on ALBERT-B and ALBERT-L. In addition, the mean values of the \textit{M-MEM} scores of all prompts for the base and large models also indicate that \textbf{model sizes} are not correlated with the memorization scores.

\subsection{Are Prompt Performances Generalizable?}

\begin{figure}
\begin{subfigure}[t]{0.25\linewidth}
\includegraphics[width=\textwidth]{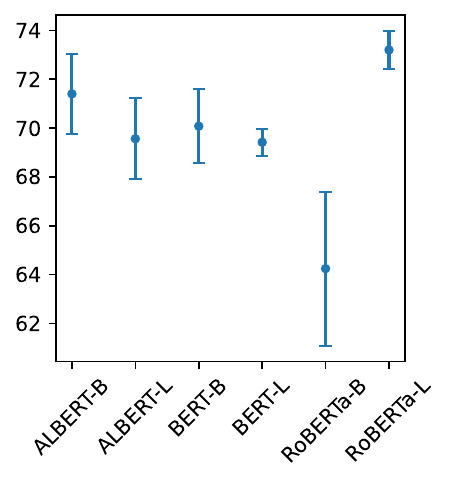}
   \caption{}
\end{subfigure}
\hfill
\begin{subfigure}[t]{0.73\linewidth}
\includegraphics[width=\linewidth]{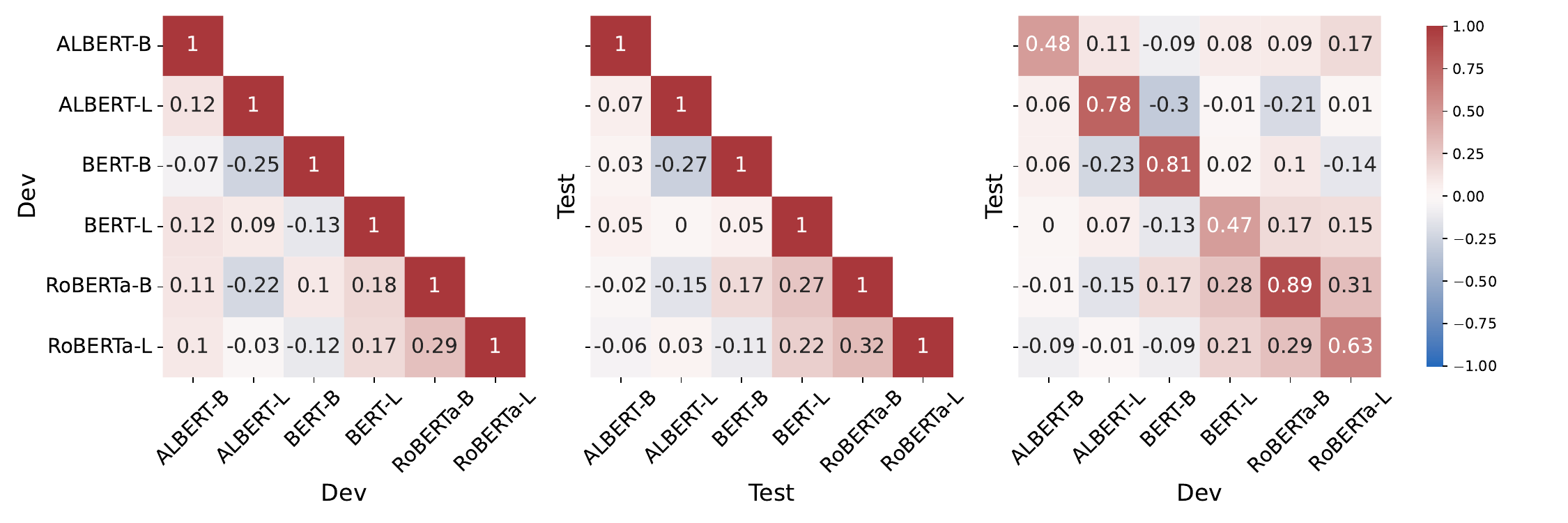}
\caption{}
\end{subfigure}
\caption{(a): Mean and standard deviation of \textit{M-MEM} scores of all prompts across models on the dev set. (b): Correlations (Kendall's $\tau$) of the \textit{M-MEM} scores of prompts across models. Left: correlations for the dev set scores. Middle: correlations for the test set scores. Right: correlations between dev and test set scores.}
\label{fig:promptRankCorrs}
\end{figure}

To examine how prompt performance generalizes across models and $PER$ pairs, we compute the correlations (Kendall's $\tau$ coefficient) among the \textit{M-MEM} scores of all the prompts for different models and data splits (Figure~\ref{fig:promptRankCorrs} (b)).
\textbf{Generalizability across models:} We observe weak correlations between prompts' \textit{M-MEM} scores for different models, even when comparing different variants of the same model on the same data split.
For example, \textit{M-MEM} scores (dev set) for BERT-B and BERT-L are negatively correlated.
From this, we conclude that the \textit{M-MEM} scores of prompts do not generalize across models---the best prompts for a given model may not be the best for other models.
\textbf{Generalizability across name sets:} Conversely, we found higher correlations between prompts' \textit{M-MEM} scores across data splits when using the same model (diagonal of the right plot in Figure~\ref{fig:promptRankCorrs}).
We conclude that the \textit{M-MEM} scores of prompts generalize to some extent across splits---for a given model, the best prompts for a given set of $PER$ are likely to work well for a different set.
Table~\ref{tab:bestWorstPrompts} illustrates this by comparing each model's best and worst prompts, showing that the (model-dependent) best and worst prompts for the dev set are still among the best and worst for the test set.
These results show that while prompt quality depends on the model, it still generalizes for different $PER$.

\subsection{What is the Best Strategy to Use the Prompt Set?}

Table \ref{tb:modified_prompt} shows the results of applying different strategies to the prompt set. \textbf{The prompt set outperforms baselines:} All the \textit{M-MEM} scores achieved by best prompts (B-PT) of the prompt set are higher than the two baselines in the dev set (only one exception in the test set), showing the necessity to detect the model memorization with diverse prompts. \textbf{Prompt engineering outperforms the prompt set:} We observe that the best/worst-modified prompts (BM-PT and WM-PT) using prompt engineering increase the score gap between the B-PT and the W-PT. The BM-PT of RoBERTa-B improves by approximately 2 percentage points up on the B-PT on the dev set, and the WM-PT decreases the \textit{M-MEM} score of BERT-L 19 percentage points. \textbf{Ensembling techniques do not bring improvement:} The ensemble techniques (EPT) do not improve/decrease the \textit{M-MEM} scores from the B-PT/W-PT scores.


\input{tables/prompt_modifying}

\subsection{Sentence Level Analysis: What Prompt Properties Impact the Performance?}

\begin{figure}[ht!]
\begin{subfigure}[t]{1.\linewidth}
\includegraphics[width=\textwidth]{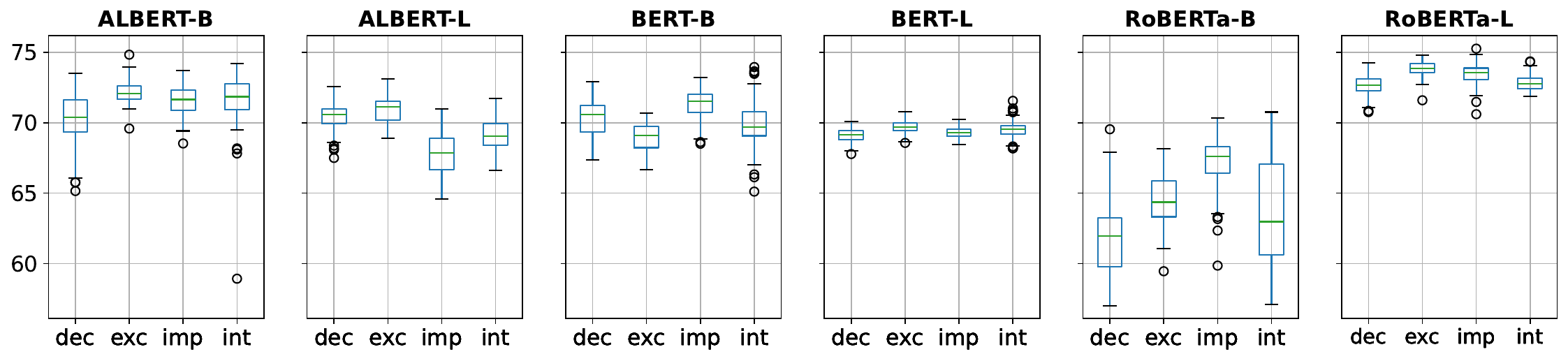}
   \caption{Prompt types (\textbf{dec}larative, \textbf{exc}lamatory, \textbf{imp}erative, and \textbf{int}errogative)}
\end{subfigure}
\hfill
\begin{subfigure}[t]{1.\linewidth}
\includegraphics[width=\linewidth]{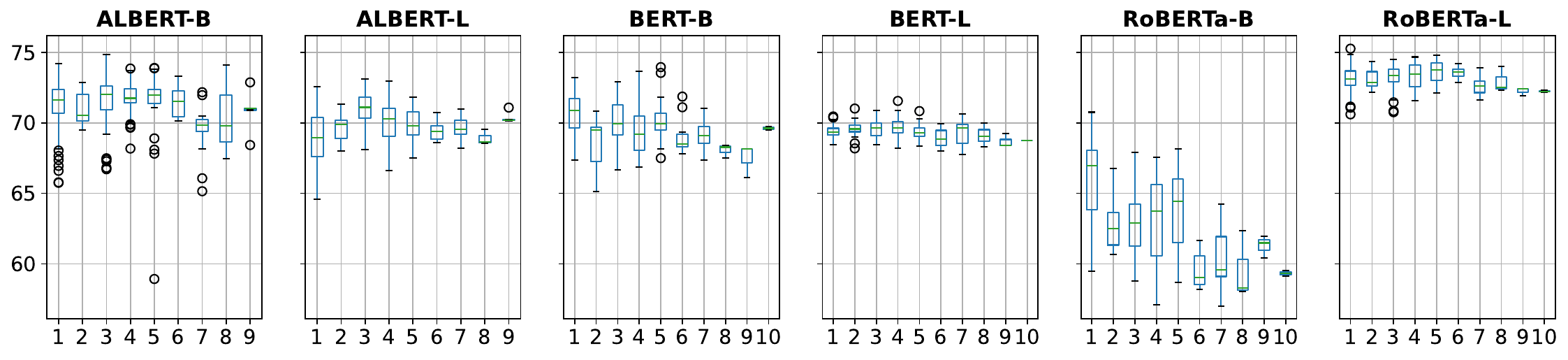}
\caption{Token positions of person's name in the prompt}
\end{subfigure}
\hfill
\begin{subfigure}[t]{1.\linewidth}
\includegraphics[width=\linewidth]{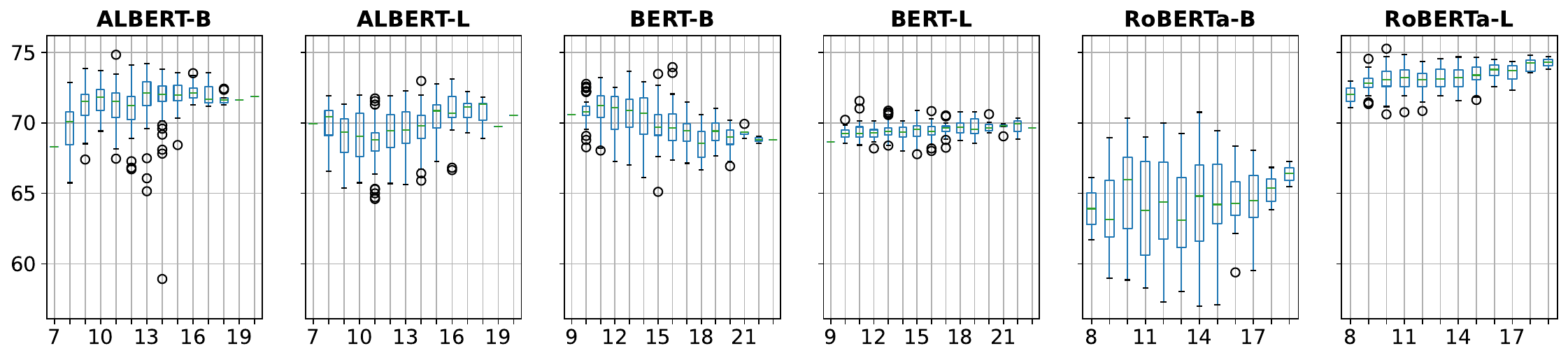}
\caption{Prompt token lengths}
\end{subfigure}
    \caption{\textit{M-MEM} scores grouped by different prompt properties on the dev set.}
    \label{fig:typeBoxplots}
\end{figure}

This section investigates different prompt properties that may influence the \textit{M-MEM} scores. 
\textbf{Prompt type:} Figure~\ref{fig:typeBoxplots} (a) presents prompts' \textit{M-MEM} scores grouped by prompt type.
While some models show similar \textit{M-MEM} scores across sentence types, others are more sensitive to particular types.
For example, RoBERTa-B has very distinct M-MEM scores for different types: imperative prompts generally yield higher scores than declarative prompts.
Conversely, BERT-L has similar \textit{M-MEM} scores profiles for different prompt types.
\textbf{Person's name token position:} Figure~\ref{fig:typeBoxplots} (b) shows prompts' \textit{M-MEM} scores grouped by the token position of the name.
The closer the name is to the beginning of the prompt, the higher the M-MEM score tends to be for RoBERTa-B and BERT-B.
\textbf{Prompt token length:} Figure~\ref{fig:typeBoxplots}~(c) contrasts prompts' \textit{M-MEM} scores and length (in number of tokens). 
We observe a weak correlation between \textit{M-MEM} scores and prompt length for most models, meaning that longer prompts are generally more effective at identifying memorization.
However, BERT-B stands out as an outlier, showing a moderate negative correlation existed between \textit{M-MEM} scores and prompt token lengths.

\subsection{Token Level Analysis: How Tokens within a Prompt Impact the Performance?} \label{aly:token}

\begin{figure}[t!]
    \centering
    \includegraphics[width=1\linewidth]{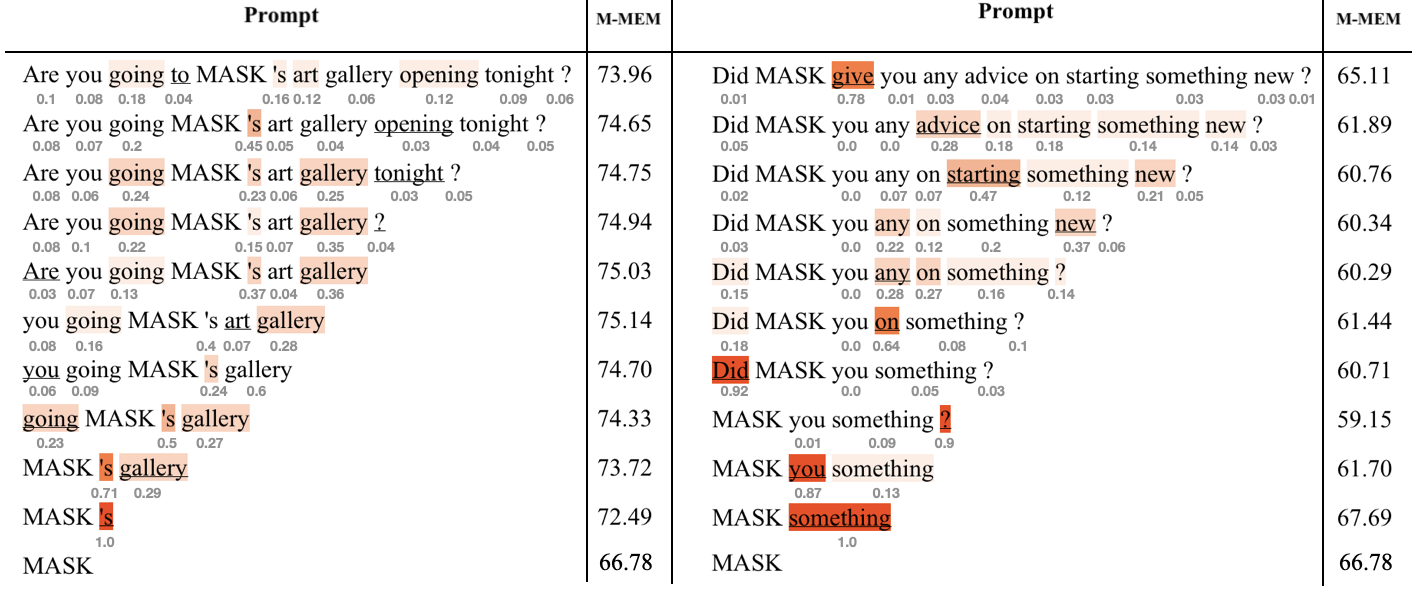}
    \caption{Analysis of the token importance in prompts and the \textit{M-MEM} scores for the BERT-B model. 
    The best-performing (left) and the worst-performing (right) prompts were selected on the dev set. 
    The heatmaps are generated with leave-one-token-out: at each step, the least important token is removed from the best-performing prompt, and the most important token is removed from the worst-performing prompt. The normalized token importance scores are under the corresponding tokens. The removed tokens are underlined.
    }
    \label{fig:tokens_removed}
\end{figure}

Figure \ref{fig:tokens_removed} shows a token-level analysis of how tokens within a prompt impact the \textit{M-MEM} scores on the example of the BERT-B model\footnote{For the analysis of other models considered in our experiments, please refer to Appendix \ref{app:token_level_analysis}.}.
Individual token importance in both the best- and worst-performing prompts (including modifications produced by token-removal) is calculated as the difference between the \textit{M-MEM} scores of the prompt with and without this token on the dev set.
These scores are then normalized using a softmax function. 
The \textit{``MASK''} token, acting as a placeholder for the $PER$, does not have an importance score.

\textbf{Removing the least important tokens from the prompt increases the \textit{M-MEM} score, while removing the most important tokens has the opposite effect.}
Iteratively removing the least important tokens from the best-performing prompt (\textit{``Are you going to MASKS's art gallery opening tonight?''}) enables an increase in the \textit{M-MEM} score up to 75.14 for the prompt \textit{``you going MASK's art gallery''}.
Continuing to remove tokens results in a slight decrease in prompt performance; however, it still outperforms the initial prompt up to the noun-phrase prompt \textit{``MASKS's gallery''} (which performs worse than the initial prompt).
In the case of the worst-performing prompt, removing the most important token \textit{give} from the initial prompt results in a major decrease in the \textit{M-MEM} score by approximately 3.5 percentage points. 
Further removal of important tokens slightly reduces the score to a minimum of 59.15 for the prompt \textit{``MASK you something ?''}. 
We observe that almost all prompts, including the ones produced by the removal of the tokens from the initial one and the initial one itself, perform worse than the prompt containing only the $PER$ name (refer to the last line in Figure \ref{fig:tokens_removed}).
The best-performing prompt consists of only one token \textit{"something"}, apart from $PER$, achieving an \textit{M-MEM} score of 67.69.

\textbf{The tokens that function as verbs and the tokens that are located near the $PER$ token tend to have a higher importance score.} Tokens like \textit{``going'', ``give'', ``practice'', ``working'', ``recommend''}, etc. (see Figure \ref{fig:tokens_removed} and Figures \ref{fig:token_importance_app_1}-\ref{fig:token_importance_app_5} in Appendix \ref{app:token_level_analysis}) have a higher influence on the \textit{M-MEM} score in each considered prompt.
A similar trend can be seen for the tokens \textit{``'s'', ``any'', ``Oh,'', ``Bravo,'', ``What,'' and ``,''} located nearer to the \textit{MASK} token (substituted by the $PER$ entity in the experiments) than tokens located further away.

\textbf{The punctuation mark at the end of each prompt has minimal influence on the prompt performance.} 
In all our experiments, the token representing the final punctuation mark consistently demonstrates a low importance score. 
On the contrary, the punctuation marks in the middle of the prompt, particularly those surrounding the \textit{MASK} token (see, for example, Figure \ref{fig:token_importance_app_1}: \textit{``Bravo, MASK, ...''}, and Figure \ref{fig:token_importance_app_2} \textit{``Oh, MASK, ...''} in Appendix \ref{app:token_level_analysis}) exhibit greater importance.
This can be explained by their role in altering sentence structure and emphasizing distinct elements within it.



\subsection{Self-attention Analysis: How Models Behave to Different Prompts and \textit{PER}?}\label{alyn: self-attention}

\begin{figure}[h!]
    \centering
    \includegraphics[width=0.95\linewidth]{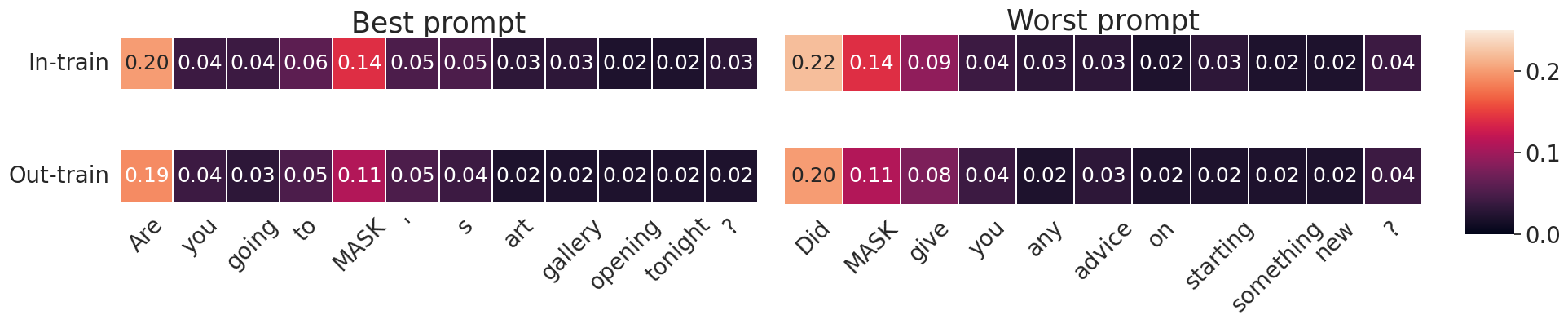}
    \caption{Attention heatmaps of the best and the worst prompts of BERT-B on the dev set averaged over all attention heads and layers. The attention weights corresponding to each prompt's ``MASK'' token are used and averaged over the In-train and Out-train $PER$s separately.}
\label{fig:att_heatmaps}
\end{figure}

An essential component of the transformer's architecture is the multi-head self-attention mechanism \citep{Vaswani2017}. 
The self-attention weights of a model 
can show us where the model attends in the sequence 
and provide insights into what affected the predictions. 
Figure~\ref{fig:att_heatmaps} shows the attention heatmaps of BERT-B for the best and the worst prompts with each prompt completed with In-train and Out-train $PER$s separately. Specifically, we first averaged the extracted attention weights of the tokens that correspond to the $PER$ (``MASK'') for all completed sentences in a group (e.g., the best prompt completed separately with all In-train $PER$s). Then we create a heatmap using the averaged values over the attention heads and the layers. Generally, we notice that attention tends to be distributed similarly given a prompt, focusing on the first and the $PER$ tokens. This observation applies to all analyzed models  (heatmaps for the other 5 models are shown in Figure \ref{fig:all_heatmaps} in Appendix~\ref{other_heatmaps}). 

\textbf{Name level analysis:}
We observe a slightly higher focus on In-train $PER$s than Out-train $PER$s when comparing the averaged attention weights on both the best and worst prompts. 
\textbf{Prompt level analysis:}
When observing the heatmaps in the prompt level, we notice that models tend to focus more (or equally for BERT-B shown in Figure \ref{fig:att_heatmaps}) on the $PER$ tokens in the best prompt than the worst, except when the $PER$ tokens in the worst prompt are positioned in the beginning the prompt.

%% file: tables/best_worst_prompt.tex
\begin{table}[!t]
    \centering
    \scriptsize
    \begin{tabular}{l|l|c|c|c|c}
    \hline
        ~ & \multirow{2}{*}{ Prompts} & \multicolumn{2}{c}{dev }  & \multicolumn{2}{c}{test} \\ 
        \cline{3-6}
        ~ & & Rank & \textit{M-MEM}   &  Rank &  \textit{M-MEM} \\ \hline
        \multirow{2}{*}{ALBERT-B} & Bravo, MASK, what an impressive performance!  & 1 &74.84& 2& 75.10 \\ 
          &Are you going to MASK's art gallery opening tonight? & -1 & 58.92&  -1& 58.74 \\ \hline

          \multirow{2}{*}{ALBERT-L} & Oh, MASK, you're a true gem in our team. & 1 & 73.12  & 3 & 71.38  \\
          & MASK, practice forgiveness towards yourself and others. & -1 & 64.60 &  -3 & 64.70  \\ \hline
          
        \multirow{2}{*}{BERT-B} & Are you going to MASK's art gallery opening tonight? & 1 &73.96& 1 & 71.84 \\ 
         & Did MASK give you any advice on starting something new?  & -1 &65.11&  -2& 63.30 \\ \hline
        
        \multirow{2}{*}{BERT-L} & What project is MASK working on? & 1 &71.56 & 2& 69.30\\ 
        & I had a chance to meet MASK's family. & -1 &67.78  &-1 & 64.42 \\ \hline

        \multirow{2}{*}{RoBERTa-B} & MASK, can you recommend a good restaurant in town? & 1 & 70.76  &2 & 70.78\\ 
        &I had a great conversation with MASK at the party. & -1 & 57.00 &  -2 & 59.63 \\ \hline
        
        \multirow{2}{*}{RoBERTa-L} & MASK, invest in meaningful relationships. & 1 & 75.27 &  1 & 72.79  \\
        &MASK, practice playing the guitar. & -1 & 70.61 &  -96 & 70.05 \\ \hline

        
    \end{tabular}
    \caption{The \textit{M-MEM} scores and corresponding ranks of the best (Rank=1) and worst (Rank=-1) prompts from the dev set on the test set.}
    \label{tab:bestWorstPrompts}
\end{table}

%% file: tables/prompt_modifying.tex
\begin{table}[t]
    \centering
    \scriptsize 
    \begin{tabular}{ll|ll|ll|ll|ll|ll|ll}
    \hline
        ~ & &\multicolumn{2}{c}{ALBERT-B} & \multicolumn{2}{c}{ALBERT-L} &\multicolumn{2}{c}{BERT-B} & \multicolumn{2}{c}{BERT-L}&\multicolumn{2}{c}{RoBERTa-B}&\multicolumn{2}{c}{RoBERTa-L} \\ \cline{3-14}
        ~ & & dev & test & dev  & test & dev & test & dev  & test& dev & test & dev & test\\ \hline
\multirow{3}{*}{\rotatebox[origin=c]{90}{BS}}&{\textbf{$\emptyset$}-PT} & 71.11 & 72.44 &70.81& 68.91& 66.78 & 66.10 &70.16&68.11 &  66.39& 65.46 & 74.40 & 70.29 \\ 
        & {One-PT} & 70.42 & 71.18 &69.02& 67.57& 72.15 &70.31 &69.91&66.49 &  68.57 & 70.72 & 72.68 & 70.22 \\
        & {Mix-PT} &68.63 & 70.02& 67.58 & 66.98 & 70.59 & 69.27 & 69.29 & 66.56 & 66.47 & 67.29 & 72.80 & 70.42\\
        \hline

\multirow{2}{*}{\rotatebox[origin=c]{90}{OPT}}&{B-PT}   &  {74.84} &   \red{75.10}&  73.12 & {71.38} &  {73.96} &  {71.84} &  {71.56} & {69.30} &  {70.76} &  {70.78} &  {75.27} & {72.79}\\ 
        &{W-PT}   & 58.92 &  58.74 & 64.60 &64.70 & 65.11 & 63.30 & 67.78 &64.42 & 57.00 & 59.63 & 70.61 &70.05\\ \hline
        
\multirow{2}{*}{\rotatebox[origin=c]{90}{PTE}} & {BM-PT}  &  \red{75.14}  & 74.96 &  \red{73.14} & \red{71.44} & \red{75.14}& \red{73.31} &  \red{73.03}& \red{72.18} &  \red{72.68} & \red{71.59} &  \red{76.28} & \red{73.33}\\ 
        
       &{WM-PT }  &  \blue{41.92}  & \blue{44.53} &  \blue{63.76} & \blue{63.52} & \blue{59.15}& \blue{62.04} &  \blue{48.68}& \blue{49.98} &  \blue{55.29} & \blue{58.21} &  \blue{69.15} & \blue{68.52} \\ \hline
\multirow{5}{*}{\rotatebox[origin=c]{90}{EPT}}&MV  & 73.24 &74.3& 70.96&69.71 &  70.75& 69.34 & 69.91 & 67.98&  64.87& 66.30 & 73.73 & 71.05\\ 
        &AVG-C  & 72.81 &73.22& 68.21  &67.43 & 69.66 &67.98  &69.92  &67.95 & 67.76 &67.67 &73.97&71.16\\ 
        &WED-C  & 72.82 & 73.22 & 68.23 & 67.44& 69.67 &  67.99&69.92& 67.95& 67.76 & 67.66 &73.97& 71.16\\ 
        & MAX-C    & 71.53 & 73.71 & 70.65 &68.99& 73.52 & 71.40 & 69.97 & 67.31& 61.47 & 63.22& 73.69 & 71.25\\ 
        & MIN-C    & 60.12 & 59.77 & 64.74 & 65.39& 68.79 & 66.56 & 68.62 &67.94 &68.35  & 68.08 & 73.08 & 71.58\\ \hline

    \end{tabular} 
    \caption{Results of using different strategies on the prompts. One-PT and Mix-PT are baselines adapted from \citet{ali_unintended_2022}. {\red{Bold}} and {\blue{underline}} highlight the highest and lowest scores respectively. }\label{tb:modified_prompt}
\end{table}

%% file: conclusion.tex
\section{Conclusion}

We studied memorization in fine-tuned MLM-based NER models. In contrast to memorization in pre-training or auto-regressive language models, entities (person's name) occur only on the input side of the fine-tuning data, and not on the output (which would be the \textit{PER} tags), and therefore cannot be detected with generation prompts as in previous works \citep{carlini2021extracting, carlini2023quantifying}. 
For measuring memorization in this setting, we compared a fixed set of 5 hand-written prompts to automatically generated 400 prompts, and we measured how well those prompts could distinguish entities that occurred in the fine-tuning data from those that did not, using a large set of candidate name entities from Wikidata. 

Detailed analysis showed that a prompt's ability to detect memorization varies between models, but is stable between different entity sets. We find a large variability in the performance of generated prompts, and that many generated prompts significantly outperform hand-written ones. The effectiveness of prompts can be increased by removing the least important tokens even if the prompts become ungrammatical after the removal.

%% file: acknowledge.tex
\section{Acknowledgements}
We thank Vasisht Duddu and N. Asokan who dedicated time and expertise to provide valuable feedback on this paper. This research has been funded by the Vienna Science and Technology Fund (WWTF) [10.47379/VRG19008] “Knowledge-infused Deep Learning for Natural Language Processing”.

%% file: appendix_token_level_analysis.tex
\subsection{Token-level Analysis}
\begin{figure}[!ht]
    \centering
    \makebox[\textwidth]{
    \includegraphics[width=\linewidth]{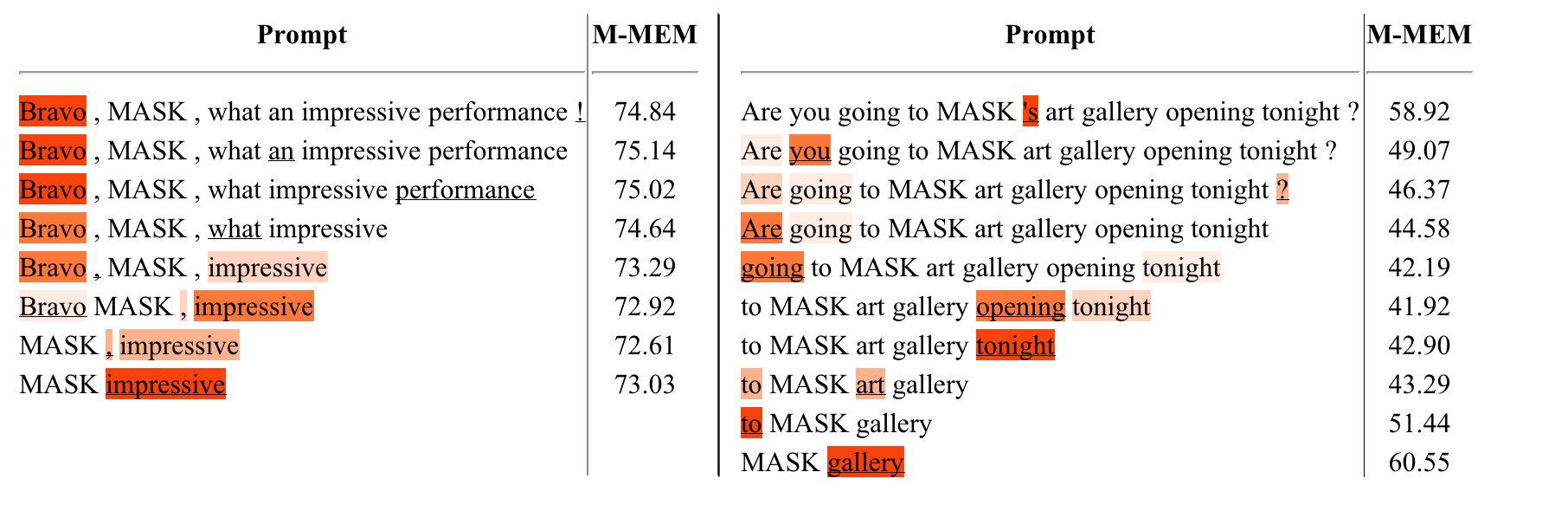}}
    \caption{Analysis of the token importance in prompts and the \textit{M-MEM} scores for the \textbf{ALBERT-B model}. 
    The best-performing (on the left) and the worst-performing (on the right) prompts were selected on the dev set. 
    The heatmaps are generated with leave-one-token-out: at each step, the least important token is removed from the best-performing prompt (left), and the most important token is removed from the worst-performing prompt (right). The removed tokens are underlined.
    }
\label{fig:token_importance_app_1}
\end{figure}

\begin{figure}[!ht]
    \centering
    \makebox[\textwidth]{
    \includegraphics[width=\linewidth]{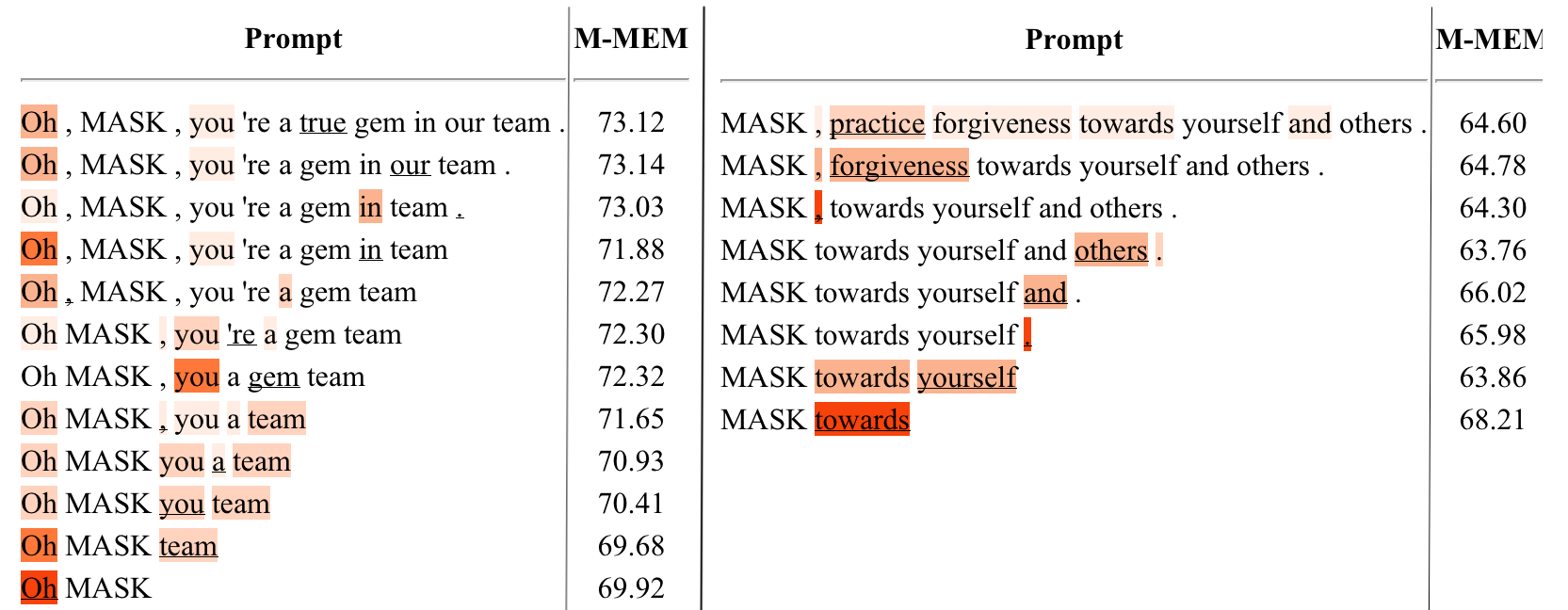}}
    \caption{Analysis of the token importance in prompts and the \textit{M-MEM} scores for the \textbf{ALBERT-L model}. 
    The best-performing (on the left) and the worst-performing (on the right) prompts were selected on the dev set. 
    The heatmaps are generated with leave-one-token-out: at each step, the least important token is removed from the best-performing prompt (left), and the most important token is removed from the worst-performing prompt (right). The removed tokens are underlined.
    }
\label{fig:token_importance_app_2}
\end{figure}

\begin{figure}[!ht]
    \centering
    \makebox[\textwidth]{
    \includegraphics[width=0.8\linewidth]{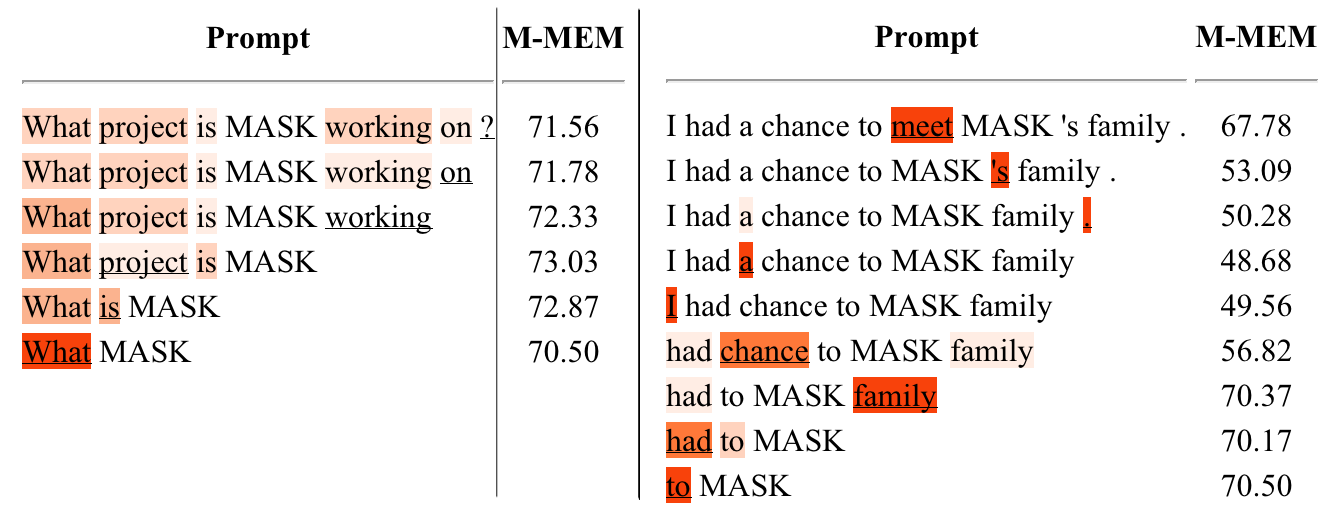}}
    \caption{Analysis of the token importance in prompts and the \textit{M-MEM} scores for the \textit{M-MEM} score of the \textbf{BERT-L model}. 
    The best-performing (on the left) and the worst-performing (on the right) prompts were selected on the dev set. 
    The heatmaps are generated with leave-one-token-out: at each step, the least important token is removed from the best-performing prompt (left), and the most important token is removed from the worst-performing prompt (right). The removed tokens are underlined.
    }
\label{fig:token_importance_app_3}
\end{figure}

\begin{figure}[!ht]
    \centering
    \makebox[\textwidth]{
    \includegraphics[width=\linewidth]{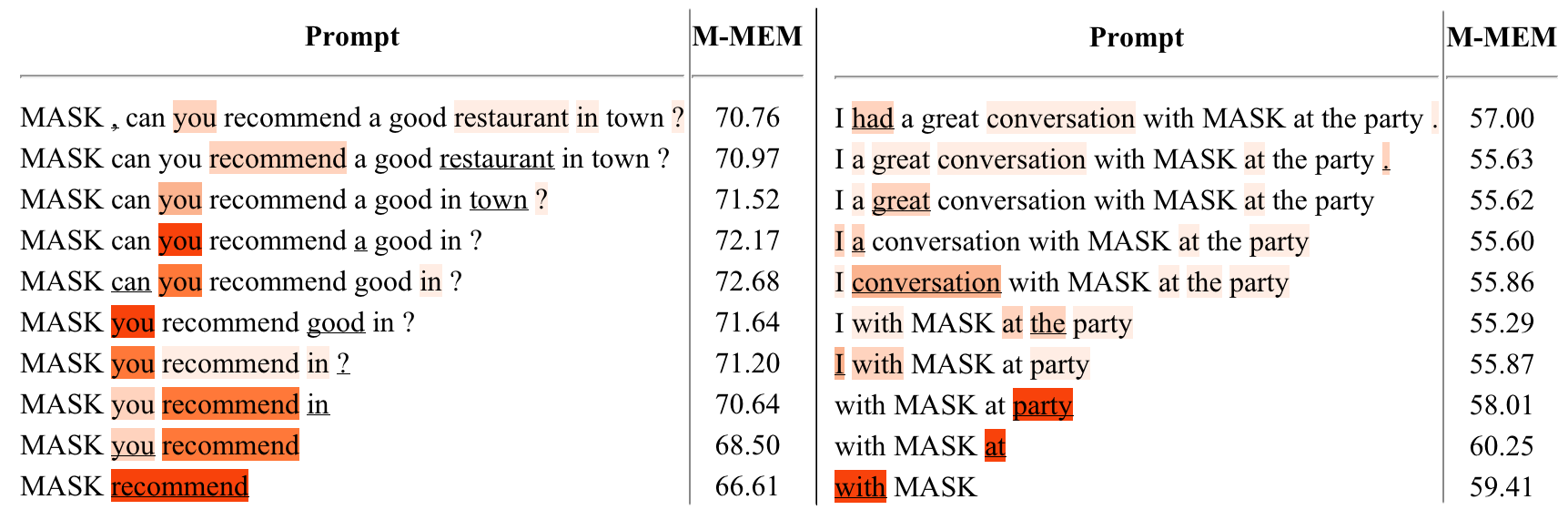}}
    \caption{Analysis of the token importance in prompts and the \textit{M-MEM} scores for the \textit{M-MEM} score of the \textbf{RoBERTa-B model}. 
    The best-performing (on the left) and the worst-performing (on the right) prompts were selected on the dev set. 
    The heatmaps are generated with leave-one-token-out: at each step, the least important token is removed from the best-performing prompt (left), and the most important token is removed from the worst-performing prompt (right). The removed tokens are underlined.
    }
\label{fig:token_importance_app_4}
\end{figure}

\begin{figure}[!ht]
    \centering
    \makebox[\textwidth]{
    \includegraphics[width=1\linewidth]{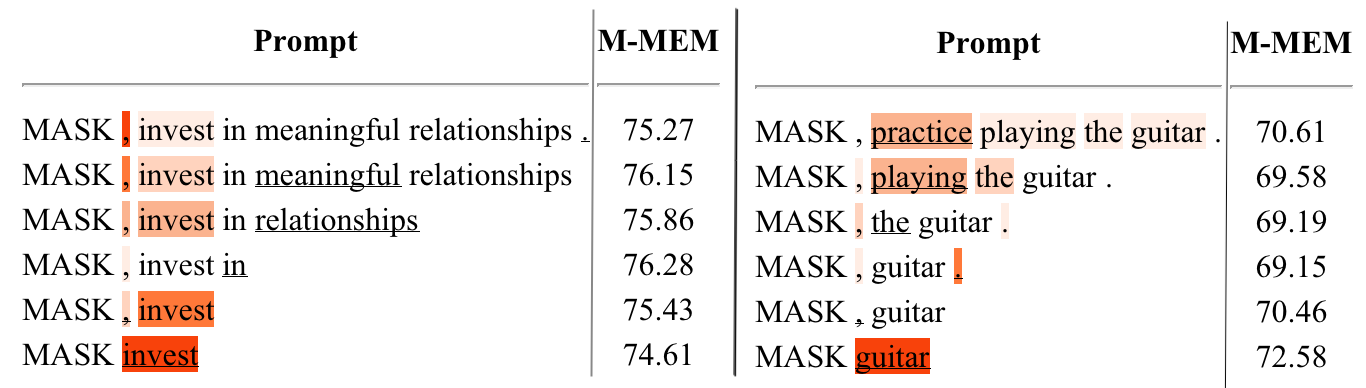}}
    \caption{Analysis of the token importance in prompts and the \textit{M-MEM} scores for the \textit{M-MEM} score of the \textbf{RoBERTa-L model}. 
    The best-performing (on the left) and the worst-performing (on the right) prompts were selected on the dev set. 
    The heatmaps are generated with leave-one-token-out: at each step, the least important token is removed from the best-performing prompt (left), and the most important token is removed from the worst-performing prompt (right). The removed tokens are underlined.
    }
\label{fig:token_importance_app_5}
\end{figure}

Same as Section \ref{aly:token}, we provide token-level analysis for the rest of the 5 models in Figure \ref{fig:token_importance_app_1}-\ref{fig:token_importance_app_5}.

\label{app:token_level_analysis}

